\documentclass[fleqn,10pt]{wlscirep}
\usepackage[utf8]{inputenc}
\usepackage[T1]{fontenc}
\usepackage[normalem]{ulem}
\usepackage{tabularx}
\usepackage[table]{xcolor}
\usepackage{geometry}
\usepackage{makecell}
\usepackage{siunitx}
\usepackage{threeparttable}
\usepackage{float}
\usepackage[justification=justified]{caption}
\usepackage{comment}
\usepackage{graphicx}
\usepackage[percent]{overpic}
\title{Towards a Generalizable AI for Materials Discovery: Validation through Immersion Coolant Screening}

\author[1,*]{Hyunseung Kim}
\author[1,*]{Dae-Woong Jeong}
\author[1]{Changyoung Park}
\author[2]{Won-Ji Lee}
\author[2]{Ha-Eun Lee}
\author[2]{Ji-Hye Lee}
\author[1]{Rodrigo Hormazabal}
\author[1]{Sung Moon Ko}
\author[1]{Sumin Lee}
\author[1]{Soorin Yim}
\author[1]{Chanhui Lee}
\author[1]{Sehui Han}
\author[2,$\dagger$]{Sang-Ho Cha}
\author[1,$\ddagger$]{Woohyung Lim}

\affil[1]{LG AI Research, Republic of Korea}
\affil[2]{Department of Chemical Engineering, Kyonggi University, Republic of Korea}

\affil[*]{These authors contributed equally.}

\affil[$\dagger$]{sanghocha@kgu.ac.kr}
\affil[$\ddagger$]{w.lim@lgresearch.ai}

\begin{abstract}
Artificial intelligence (AI) has emerged as a powerful accelerator of materials discovery, yet most existing models remain problem-specific, requiring additional data collection and retraining for each new property. Here we introduce and validate GATE (Geometrically Aligned Transfer Encoder)—a generalizable AI framework that jointly learns 34 physicochemical properties spanning thermal, electrical, mechanical, and optical domains. By aligning these properties within a shared geometric space, GATE captures cross-property correlations that reduce \textit{disjoint-property bias}\textemdash a key factor causing false positives in multi-criteria screening. To demonstrate its generalizable utility, GATE\textemdash without any problem-specific model reconfiguration\textemdash applied to the discovery of immersion cooling fluids for data centers, a stringent real-world challenge defined by the Open Compute Project (OCP). Screening billions of candidates, GATE identified 92,861 molecules as promising for practical deployment. Four were experimentally or literarily validated, showing strong agreement with wet-lab measurements and performance comparable to or exceeding a commercial coolant. These results establish GATE as a generalizable AI platform readily applicable across diverse materials discovery tasks.
\end{abstract}

\begin{document}

\flushbottom
\maketitle

\section*{Introduction}

Materials discovery increasingly relies on artificial intelligence (AI) to accelerate innovation. Recent advances have shown how AI can accelerate this process—by enabling rapid virtual screening to reduce costly synthesis and testing\cite{merchant2023scaling, gentrl, szymanski2023autonomous, burger2020mobile, coley2019robotic, liu2022experimental}, by uncovering new structure–property relationships that expand fundamental understanding\cite{tao2021machine, bradford2023chemistry, kim2025advancing, hong2022melting, zhang2024machine, tao2021benchmarking, armeli2023machine, li2024reusability, xie2025accelerating, na2020tuplewise, barnett2020designing}, and by identifying materials that have progressed to real-world deployment\cite{bradford2023chemistry, kim2025advancing, gentrl, bai2019accelerated, burger2020mobile, barnett2020designing, hong2022melting}. These successes illustrate the growing role of AI as a practical tool for guiding experimental workflows.

Despite this progress, most efforts in the field have remained confined to single-property models, which treat each property as an isolated task rather than learning their inherent correlations\cite{tao2021machine, waters2022predicting, zeng2021correlation, armeli2023machine, na2020tuplewise, tao2021benchmarking, zhang2024machine, hong2022melting}. While this single-task paradigm has yielded useful predictors, it poses inherent risks when applied to real materials development. The majority of applications demand simultaneous optimization of multiple properties—such as thermal stability, electrical insulation, and environmental compatibility—rather than a single metric. These properties are often correlated or even antagonistic, meaning that optimizing them independently ignores the underlying trade-offs\cite{xu2025multi, gurnani2024ai, zunger2018inverse, greenaway2021integrating, bai2019accelerated, kim2024materials, xie2025accelerating}.

\begin{figure}[h!]
    \centering
    \makebox[\textwidth][c]{%
        \includegraphics[width=1.1\textwidth]{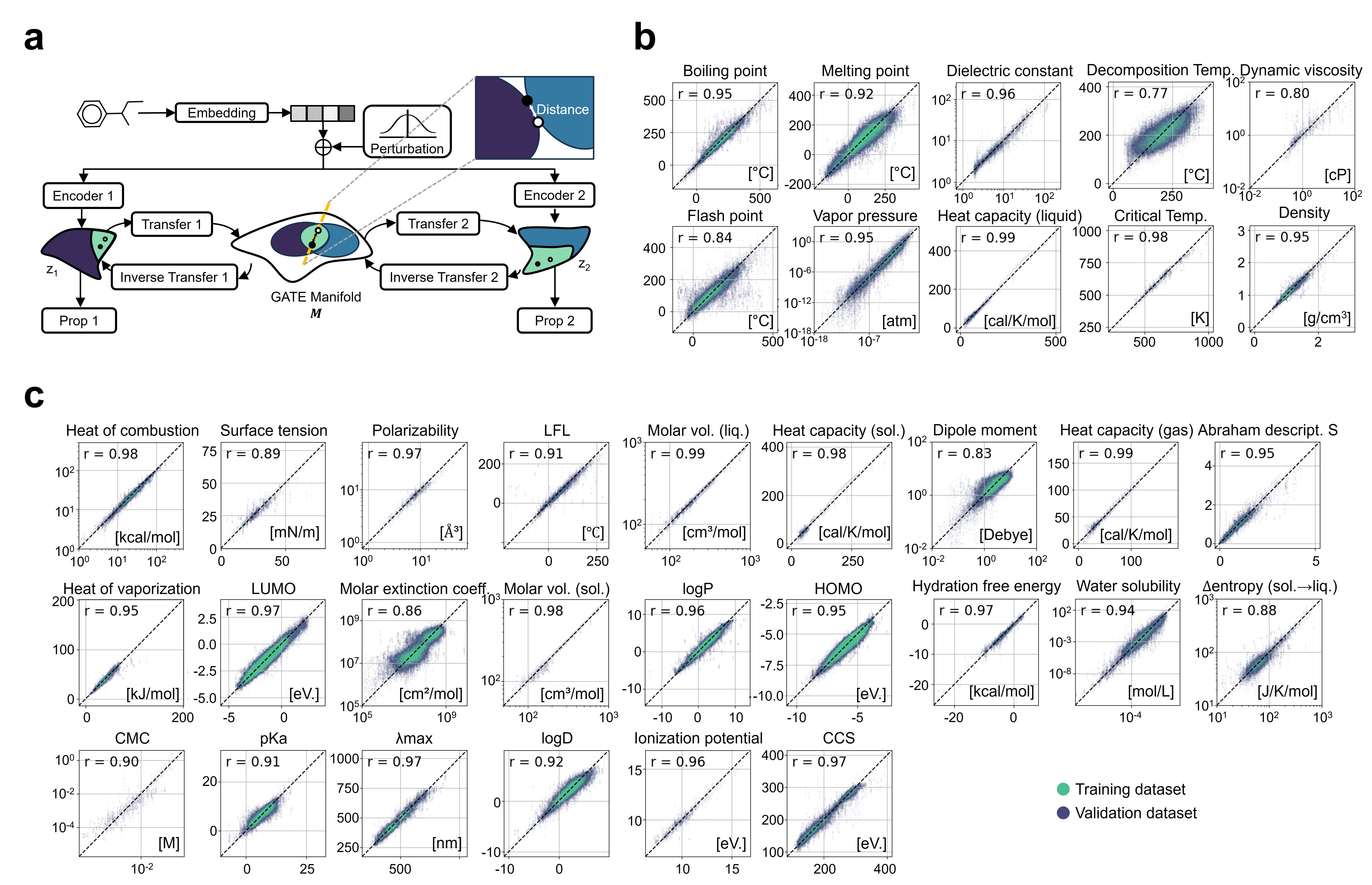}
    }
    \caption{\textbf{GATE enables generalizable molecular property prediction across diverse tasks.}  
    \textbf{a.} Schematic illustration of the GATE architecture, which learns transferable knowledge across different molecular–property relationships via a shared latent manifold.  
    \textbf{b–c.} GATE’s predictive performance across multiple physicochemical properties. Each plot compares ground-truth reference values (x-axis) with GATE-predicted values (y-axis). Panel \textbf{b} shows properties directly relevant to immersion cooling fluids for data centers, while panel \textbf{c} extends to additional properties beyond this application, highlighting GATE’s broad generalizability across 34 distinct tasks. Mint points represent training datasets and navy points represent validation datasets, with Pearson correlation coefficients ($r$) reported for the validation datasets.}
    \label{fig:gate}
\end{figure}

Consequently, when independently predicted properties are combined to satisfy multiple design criteria, a systematic bias can arise. In this case, compounds that do not truly meet the requirements but are predicted to pass all thresholds (false positives) are more easily retained than compounds whose predictions and actual values both meet the criteria (true positives). We term this \textit{disjoint-property bias}. It occurs because each property is modeled in isolation, so correlations and trade-offs that naturally emerge from molecular structure are neglected. As the number of criteria increases, the bias becomes more pronounced, yielding candidates that appear promising in silico yet fail experimental validation (Methods and Supplementary Fig.~1).

We introduce the Geometrically Aligned Transfer Encoder (GATE)\cite{gate1, gate2, gate3, gate4}, a general framework that jointly learns 34 physicochemical properties across thermal, electrical, mechanical, and optical domains (see Fig.~\ref{fig:gate}). GATE aligns molecular representations across tasks so that knowledge learned from well-characterized properties can improve the learning of other properties even with sparse or noisy data. In this process, GATE explicitly learns correlations among properties, thereby mitigating \textit{disjoint-property bias} and enabling more reliable multi-criteria materials discovery.

The selection of these 34 properties was guided by extensive consultation with domain experts across diverse materials development—including polymers, electronic materials, batteries, displays, cosmetics, and agricultural chemicals, as well as functional additives. Each participating group identified physicochemical parameters most critical to their respective design and optimization workflows, which were consolidated to define a property space with the highest cross-domain utility. Consequently, GATE encompasses thermal, electrical, mechanical, and interfacial properties that collectively underpin a wide spectrum of real-world materials discovery problems. By training on this expert-curated property set, GATE functions as a generalizable AI platform that can be directly deployed across heterogeneous design tasks without retraining or task-specific modification, providing a unified foundation for accelerating diverse materials development processes across domains.

To validate practicality, we directly applied GATE\textemdash without any problem-specific model reconfiguration\textemdash to a real-world materials design problem: the discovery of immersion cooling fluids for data centers. Candidate fluids are guided by diverse physicochemical constraints—spanning thermal, mechanical, electrical, and environmental requirements—codified in detail by the Open Compute Project (OCP) guidelines\cite{ocp2022immersion}. Using GATE, billions of virtual and purchasable compounds were screened, promising candidates were identified, and their properties were validated through experimental measurements (thermogravimetric analysis, TGA; differential scanning calorimetry, DSC) and literature values.

Beyond this demonstration, GATE exhibits strong generalization in data-scarce and out-of-distribution (OOD) regimes by leveraging correlations among multiple properties. Such regimes are particularly relevant to materials discovery, as promising candidates often lie outside the range of existing data due to their extreme or unconventional physicochemical characteristics. By transferring knowledge across properties, GATE reduces false positives that arise from \textit{disjoint-property bias}, thereby improving the reliability of multi-criteria predictions and enhancing its practical utility for real-world materials design.

Together, these findings establish GATE as a ready-to-use and generalizable AI model capable of discovering viable materials under realistic design constraints, and highlight the importance of learning cross-property relationships for robust and trustworthy materials discovery.

\section*{Results}
\subsection*{Immersion Cooling Fluids for Data Centers}

\begin{table}[b!]
\centering
\begin{threeparttable}
\makebox[\textwidth][c]{%
\begin{minipage}{\textwidth}
\caption{\label{tab:ocp} Specification of OCP for single-phase immersion cooling fluids and target specification used in this paper.}

\small
\begin{tabular}{|l|l|l|l|}
\hline
\multicolumn{2}{|c|}{\textbf{OCP Guide Specification\cite{ocp2022immersion}}} & \multicolumn{2}{c|}{\textbf{Criteria (This Paper)}} \\
\hline
\textbf{Parameter} & \textbf{Spec Value} & \textbf{GATE Parameter} & \textbf{Spec Value} \\
\hline
\multicolumn{4}{|l|}{\textbf{Thermal Specifications}} \\
\hline
Boiling Point & 
\makecell[l]{> 155~\textcelsius{} @1atm, 1500m;\\ > 150~\textcelsius{} @1atm, sea level} & 
Boiling Point & 
> 150~\textcelsius{} @1atm, sea level \\
\hline
Pour Point & < -30~\textcelsius{} & Melting Point & < -30~\textcelsius{} \\
\hline
Flash Point (Closed Cup) & 
\makecell[l]{> 155~\textcelsius{} @1atm, 1500m;\\ > 150~\textcelsius{} @1atm, sea level} & 
Flash Point & 
> 140~\textcelsius{} @1atm, sea level \\
\hline
Critical Temperature & > 155~\textcelsius{} & Critical Temperature & > 155~\textcelsius{} \\
\hline
Auto Ignition Temperature & Refer to IEC 62368-1 & & \\
\hline
 & & Decomposition Temperature & > 150~\textcelsius{} @1atm \\
\hline
 & & Specific Heat & Preferably high \\
\hline
\multicolumn{4}{|l|}{\textbf{Pressure Specifications}} \\
\hline
Critical Pressure & ASTM D6378 & & \\
\hline
Vapor Pressure & < 0.8~kPa & Vapor Pressure & < 0.8~kPa \\
\hline
\multicolumn{4}{|l|}{\textbf{Figures of Merit}} \\
\hline
\makecell[l]{\tnote{1}\hspace{1.5mm}FOM1 (Natural Convection)} & > 45 (Tier2), > 35 (Tier1) & & \\
\hline
\makecell[l]{\tnote{2}\hspace{1.5mm}FOM2 (Developing Laminar Flow)} & > 19 & & \\
\hline
\makecell[l]{\tnote{3}\hspace{1.5mm}FOM3 (Dynamic Viscosity)} & < 0.015~N·s/m\textsuperscript{2} & Dynamic Viscosity & < 0.015~N·s/m\textsuperscript{2} \\
\hline
Density & < 2,000~kg/m\textsuperscript{3} & Density & < 2,000~kg/m\textsuperscript{3} \\
\hline
\multicolumn{4}{|l|}{\textbf{Electrical Parameters}} \\
\hline
Dielectric Strength over lifetime & > 6~kV/mm & & \\
\hline
\makecell[l]{Dielectric Constant \\
@ 20 MHz -- 40 GHz} & $\leq$ 2.3 & 
\makecell[l]{Dielectric Constant\\@ 1 kHz} & $\leq$ 6 \\
\hline
\makecell[l]{Dielectric Loss Tangent \\
@ 20 MHz -- 40GHz} & $\leq$ 0.05 & & \\
\hline
Volume Resistivity & > 1.0e11~$\Omega\cdot$cm & & \\
\hline
\multicolumn{4}{|l|}{\textbf{Environmental Parameters}} \\
\hline
Ozone Depletion Potential & 0 & \makecell[l]{Fluorine, Chlorine, Bromine} & Not Contained \\
\hline
\multicolumn{4}{|l|}{\textbf{Others}} \\
\hline
Color & Preferably transparent & Sulfur & Not Contained \\
\hline
Toxicity & Preferably non-toxic & \makecell[l]{Fluorine, Chlorine, Bromine, \\ Iodine, Aromatic ring} & Not Contained \\
\hline
Odor & Preferably odorless & Aromatic ring & Not Contained \\
\hline
\end{tabular}

\begin{tablenotes}
\footnotesize
\item[1] FOM1 refers to a score for natural convection in heatsink, calculated as $k\left(\frac{\beta c_p\rho^2}{\mu k}\right)^{0.2813}$.
\item[2] FOM2 refers to a score for heat transfer in laminar flow, calculated as $k\frac{\rho c_p}{\mu}$.
\item[3] FOM3 refers to dynamic viscosity in N·s/m\textsuperscript{2}.
\item[] Where $k$ is thermal conductivity [$W/m\cdot K$], $\beta$ is thermal expansion coefficient [$1/K$], $\rho$ is density [$kg/m^3$], $c_p$ is specific heat [$J/g\cdot$\textcelsius{}], and $\mu$ is dynamic viscosity [$N \cdot s/m^2$].
\end{tablenotes}

\end{minipage}
}
\end{threeparttable}
\end{table}

The rapid rise of artificial intelligence has greatly increased heat loads in modern data centers, creating an urgent need
for advanced cooling solutions. Immersion cooling—where IT equipment is submerged in a thermally conductive
yet electrically insulating liquid—has emerged as a promising strategy to address this challenge. Viable immersion
fluids must simultaneously satisfy a wide range of physicochemical requirements, including thermal stability, fluidity,
electrical insulation, and environmental safety. These specifications are codified in detail by OCP, making immersion
cooling a real-world challenge for validating AI-driven materials discovery (Table~\ref{tab:ocp}, left). This provides a natural setting to examine \textit{disjoint-property bias}, which becomes more pronounced as the number of criteria increases. Because the OCP guideline specifies a wide range of physicochemical requirements with clear definitions,
they serve as an appropriate benchmark to test whether GATE can support reliable multi-criteria discovery.

Among the 34 physicochemical properties supported by GATE, ten properties\textemdash boiling point, melting point, flash point, critical temperature, decomposition temperature, specific heat, vapor pressure, dynamic viscosity, density, and dielectric constant\textemdash are directly relevant to immersion cooling fluids and collectively cover the key requirements specified by OCP (Table~\ref{tab:ocp}). This broad coverage allows GATE to be applied directly to the immersion cooling problem without a problem-specific framework reconfiguration.

To apply the OCP guidelines using GATE, we introduced four pragmatic adaptations (Table~\ref{tab:ocp}, right):
(i) a dielectric constant threshold of $\leq 6$ at 1 kHz, reflecting that dielectric constants decrease with frequency while OCP specifies $\leq 2.3$ in the 20–40 MHz range (see Supplementary Fig.~2);
(ii) a flash-point threshold of 140 °C, reflecting higher variance and lower sensitivity of model predictions around the threshold region;
(iii) substitution of melting point for pour point, ensuring fluidity down to $-30$~\textcelsius{} since GATE does not provide pour point estimates; and
(iv) adaptation of the “Others” criteria—color, toxicity, and odor—to exclude compounds containing halogens (F, Cl, Br, I) or aromatic rings.

\begin{figure}[ht!]
    \centering
    \makebox[\textwidth][c]{%
        \includegraphics[width=1.1\textwidth]{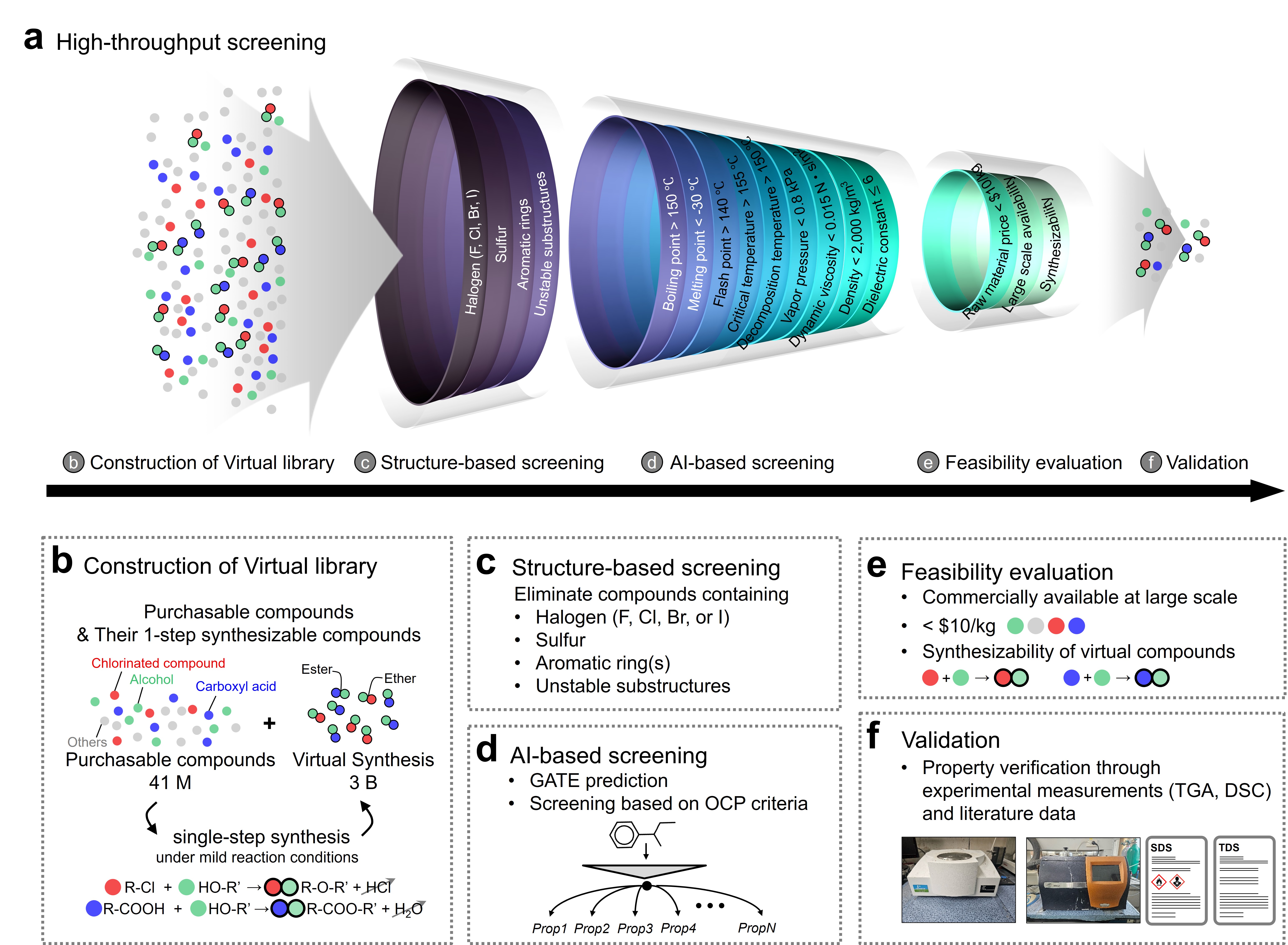}
    } 
    \caption{\textbf{High-throughput screening workflow for immersion cooling fluids.}  
    \textbf{a} Overall pipeline integrating virtual library construction, structure-based filtering, AI-based property prediction, feasibility evaluation, and final validation.  
    \textbf{b} Virtual library: purchasable compounds and single-step synthesizable derivatives.  
    \textbf{c} Structure-based filtering: removal of compounds with halogens, sulfur, aromatic rings, or unstable substructures.  
    \textbf{d} AI-based screening: GATE predictions benchmarked against OCP criteria.  
    \textbf{e} Feasibility: evaluation of cost, large-scale availability, and synthetic accessibility.  
    \textbf{f} Validation: property verification through experiments (TGA, DSC) and literature values.}
    \label{fig:workflow}
\end{figure}

\subsection*{AI-enabled high-throughput screening workflow}

To accelerate the discovery of immersion cooling fluids, we established an AI-enabled high-throughput screening workflow (Fig.~\ref{fig:workflow}). The process integrates five key stages, from virtual library construction to experimental and literature validation, thereby linking large-scale computational screening with practical materials discovery.

The workflow begins with the construction of a large virtual library, composed of purchasable compounds and their virtually synthesized derivatives generated under mild, single-step reaction conditions (Fig.~\ref{fig:workflow}b). Compounds containing undesirable elements or unstable substructures are subsequently removed through structure-based pre-screening (Fig.~\ref{fig:workflow}c). AI-based screening is then applied to predict the physicochemical properties of candidate molecules using GATE, benchmarked against the adapted criteria defined in Table~\ref{tab:ocp} (right) (Fig.~\ref{fig:workflow}d). Shortlisted compounds then undergo feasibility evaluation, considering commercial availability, cost, and synthetic accessibility (Fig.~\ref{fig:workflow}e). Finally, experimental and literature validation confirm the physicochemical properties of selected candidates (Fig.~\ref{fig:workflow}f), ensuring their suitability as single-phase immersion cooling fluids.

\subsection*{Characterization and Validation of Screened Candidates}

\begin{figure}[htbp]
    \centering
    \makebox[\textwidth][c]{%
        \begin{overpic}[width=1.1\textwidth]{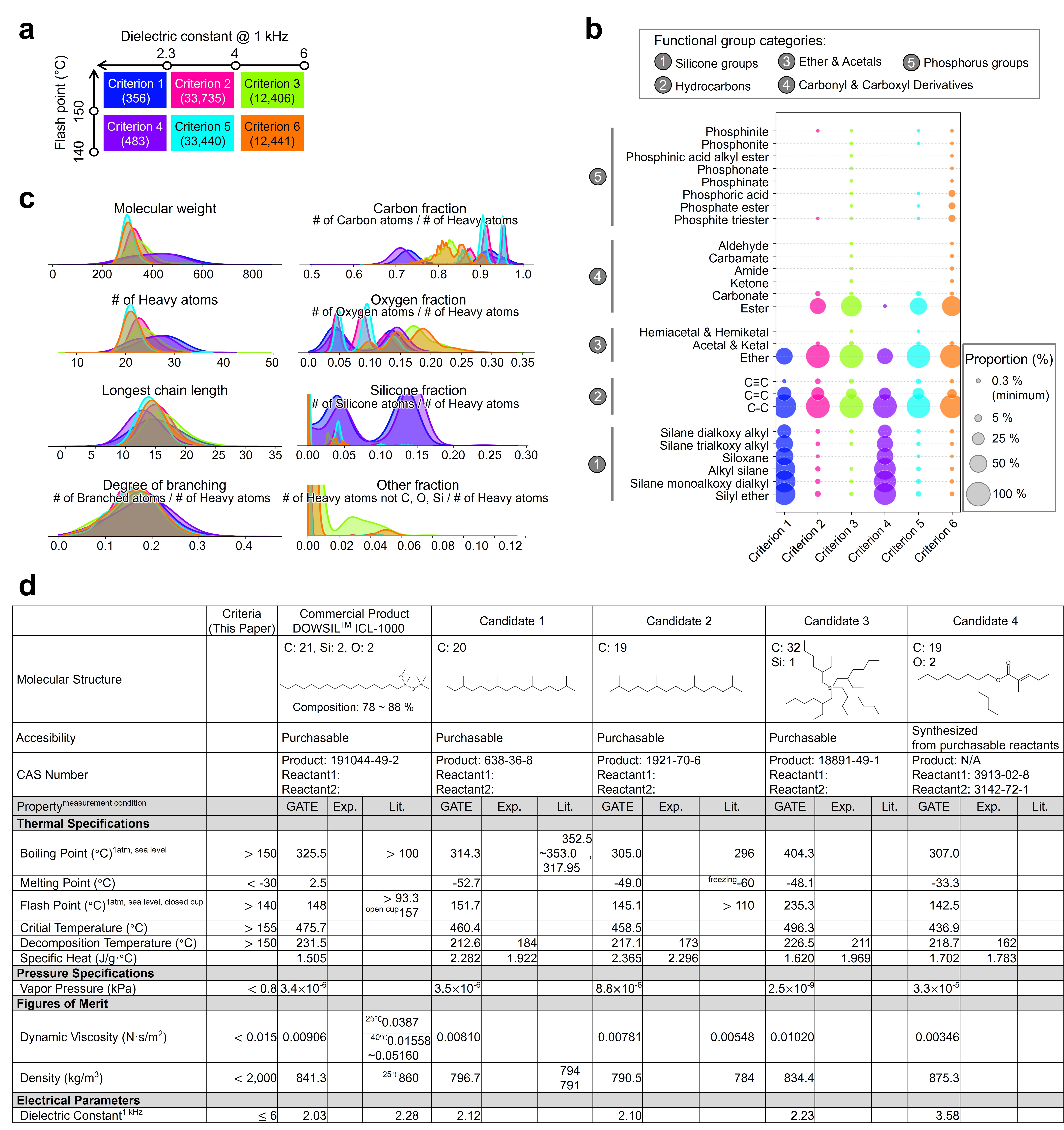}
            \put(35.8,44.0){\scriptsize\cite{DowSDS2025}} 
            \put(37.1,24.1){\scriptsize\cite{DowSDS2025}} 
            \put(37.1,20){\scriptsize\cite{DowSDS2025}} 
            \put(37.1,18.7){\scriptsize\cite{DowSDS2025}} 
            \put(37.1,9){\scriptsize\cite{DowSDS2025}} 
            \put(37.1,6){\scriptsize\cite{DowSDS2025}} 
            \put(37.1,4.1){\scriptsize\cite{DowSDS2025}} 
            \put(37.1,0.7){\scriptsize\cite{DowSDS2025}} 
            \put(51.1,23.9){\scriptsize\cite{SciFinder}} 
            \put(51.3,22.8){\scriptsize\cite{SigmaAldrichSDSPhytane}} 
            \put(51.3,4.7){\scriptsize\cite{SciFinder}} 
            \put(51.3,3.5){\scriptsize\cite{SigmaAldrichSDSPhytane}} 
            \put(66.7,24.14){\scriptsize\cite{SciFinder}} 
            \put(66.7,21.33){\scriptsize\cite{Hawley2016}} 
            \put(66.7,19.55){\scriptsize\cite{SigmaAldrichSDSPristane}} 
            \put(66.7,7.85){\scriptsize\cite{SciFinder}} 
            \put(66.7,4.1){\scriptsize\cite{SciFinder}} 
        \end{overpic}
    }
    \caption{\textbf{Characterization of screened candidates for immersion cooling fluids.}
    \textbf{a.} Definition of screening criteria based on dielectric constant and flash point thresholds. Colors denote the six criteria used for downstream analyses (panels b–c). Numbers in parentheses indicate the count of compounds meeting each criterion, while all other OCP requirements were held constant.  
    \textbf{b.} Distributions of compounds containing each functional group across screening criteria sets.  
    \textbf{c.} Structural features of candidates, including molecular weight, chain topology, and heavy atom composition.  
    \textbf{d.} Comparison of GATE predictions (GATE), experimental measurements (Exp.), and literature values (Lit.) for commercial fluids and screened candidates. Measurement conditions shown next to each property name indicate the default criteria, while values obtained under different conditions are marked with superscript.}
    \label{fig:screened}   
\end{figure}

We analyzed the candidate fluids emerging from the workflow to understand their chemistries, structures, and properties (Fig.~\ref{fig:screened}). As mentioned previously, since the OCP guideline defines dielectric constant $\leq 2.3$ over 20~MHz–40~GHz while GATE predictions are provided at 1~kHz, we adopted a pragmatic screening threshold of dielectric constant $\leq 6$ at 1~kHz. This choice is motivated by the well-known decrease of dielectric constant with frequency and is illustrated by literature data for neoprene redrawn in Supplementary Fig.~2. As this adaptation is heuristic rather than a strict derivation, we further partitioned the candidates into sub-criteria to analyze the characteristic features of each group, thereby providing insights into the molecular traits associated with meeting different performance thresholds.

We therefore partitioned candidates by dielectric constant ($\leq$2.3, 2.3--4, 4--6 at 1 kHz) and flash point (140--150 \textcelsius{}, $>$150 \textcelsius{}), while holding all other requirements to the criteria (Table~\ref{tab:ocp}, right). Fig.~\ref{fig:screened}a illustrates the thresholds of these partitions and the number of compounds identified under each criterion. Across the 41 million purchasable compounds and 2.54 billion in-silico reaction products examined, this screening yielded 356 candidates satisfying \textit{Criterion 1}, and a total of 92,861 compounds meeting at least one of \textit{Criteria 1–6}. Within the purchasable set alone, the corresponding numbers were 6 and 46, underscoring that the majority of viable candidates emerged from the virtual reaction space.

Functional group analysis (Fig.~\ref{fig:screened}b) reveals clear compositional patterns: low-dielectric sets (\textit{Criteria 1} and \textit{4}) are enriched in silicone-containing groups, consistent with their recognized insulating behavior, whereas their prevalence drops markedly in higher dielectric constant sets (\textit{Criteria 2, 3, 5, 6}). Broader trends include frequent occurrence of hydrocarbons, ethers/acetals, carbonyl \& carboxyl derivatives, and phosphorus groups; minor contributions from alcohols, peroxides, nitrogen/boron species, and metal–organic alkoxides appear mainly in higher dielectric constant sets. The similarity observed among the sets (1 with 4, 2 with 5, and 3 with 6) aligns with the dielectric constant partitions, indicating that functional group composition is governed primarily by dielectric constant rather than flash point.

Structural descriptors (Fig.~\ref{fig:screened}c)—molecular weight, heavy-atom count, longest-chain length, degree of branching, and elemental fractions (C/O/Si/others)—capture both topological and compositional patterns. Here, heavy atoms indicate atoms excluding hydrogen; the longest-chain length is the number of heavy atoms in the longest continuous chain; and the degree of branching is the fraction of heavy atoms connected to three or more neighbors.

Candidates in \textit{Criterion 1} (best-performing set) exhibit median values of molecular weight--419, 27 heavy atoms, a longest chain length of 15, and a branching degree of 0.18. Their compositions are confined to C, O, and Si with fractions 0.88, 0.08, and 0.05, respectively, and an \emph{other} fraction of 0. The C/O/Si distributions are clearly bimodal, separating silicone-containing molecules from silicone-free, carbon-dominated molecules; oxygen appears in both classes but is not obligatory.

Across criteria, topology-type descriptors (Fig.~\ref{fig:screened}c, left) cluster as (1 with 4), (2 with 5), and (3 with 6), whereas elemental compositions (Fig.~\ref{fig:screened}d, right) cluster as (1 with 4), (2 with 3) and (5 with 6). These patterns suggest that flash point is linked more strongly to molecular topology, while dielectric constant is governed primarily by elemental composition—consistent with its dependence on molecular polarity.

Feasibility evaluation (Fig.~\ref{fig:workflow}d) was then applied to the 92,861 GATE-identified candidates, considering factors such as commercial availability in ton-scale quantities, price below \$10 per kg, and synthetic accessibility. Because verifying suppliers and prices for all candidates was time-consuming and impractical, we examined 313 compounds, of which only 8 (2.6\%) met the commercial availability requirement. This limited fraction likely reflects market dynamics rather than fundamental synthetic barriers, and the true number of accessible candidates may be higher.

From these, we selected four final candidates for experimental verification. Candidates 1–3 were directly purchasable and thus tested without synthesis, whereas Candidate 4 originated from the virtual reaction library and was obtained through a targeted laboratory synthesis, thereby confirming the feasibility of incorporating virtually designed compounds into the screening pipeline

These candidates, together with a commercial immersion fluid, were benchmarked by comparing GATE predictions (GATE), experimental measurements (Exp.), and literature values (Lit.) across key physicochemical properties. Decomposition temperature was measured using thermogravimetric analysis (TGA, onset point) and specific heat using differential scanning calorimetry (DSC, 20\textcelsius{}). For other properties that could not be measured with our available instrumentation, we surveyed the literature and incorporated reported values into Fig.~\ref{fig:screened}d. 

Overall, the four candidates identified through GATE exhibited performance competitive with the commercial benchmark, and in some cases even surpassed it based on predicted properties. Notably, Candidate 3 was predicted to have a substantially higher flash point than the commercial product, while viscosity values for most candidates were lower than those of the benchmark.

Notably, many screened compounds are structurally related to mineral/synthetic oils and to natural/synthetic esters. These classes were highlighted as promising in the OCP white paper\cite{ocp2022compatibility}, underscoring the reliability of our AI-driven discovery pipeline. The agreement among GATE predictions (GATE), experimental measurements (Exp.), and literature values (Lit.) across key physicochemical properties further supports the robustness of the approach. Beyond reproducing known classes, the screening also revealed less conventional motifs—identified specifically in the context of immersion cooling fluids for data centers—including phosphorus groups, alcohols, peroxides, nitrogen/boron species, and metal–organic alkoxides. Although the practical viability of these minor classes remains to be validated, their emergence highlights the potential of AI-guided discovery to expand the design space beyond human-established heuristics.

\subsection*{Benchmarking GATE Performance under Out-of-Distribution Conditions}

\begin{figure}[t!]
    \centering
    \makebox[\textwidth][c]{%
        \includegraphics[width=1.1\textwidth]{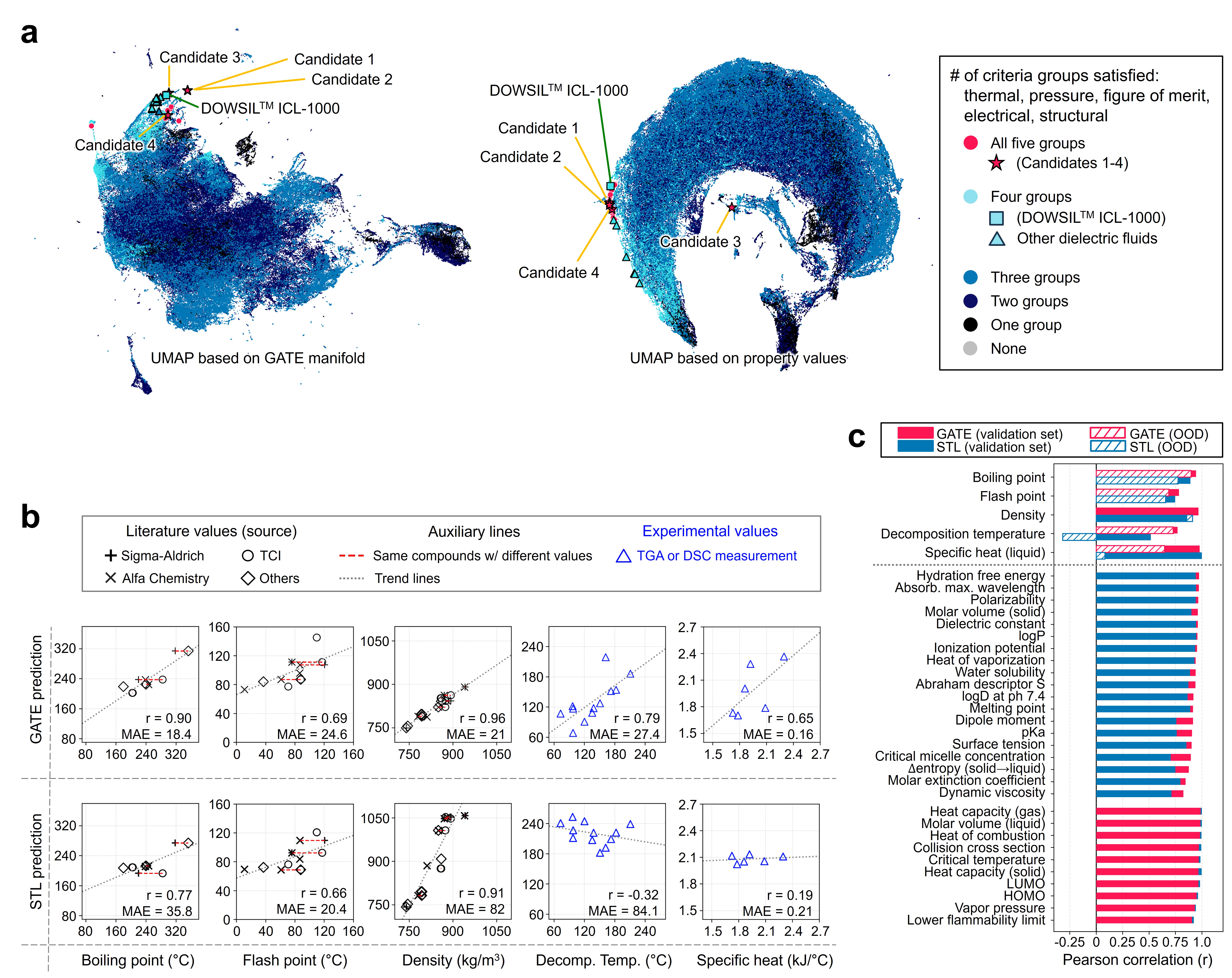}
    }
    \caption{\textbf{Benchmarking GATE and STL performance under out-of-distribution (OOD) conditions.}
    \textbf{a.} UMAP visualizations of the GATE manifold and immersion fluids. Left: GATE manifold. Right: ten key physicochemical properties relevant to immersion cooling. Both maps show that immersion fluids and screened candidates lie in OOD regions relative to the GATE training domain.
    \textbf{b.} Comparison between AI-predicted and reference (experimental or literature) values for five benchmarked properties—boiling point, flash point, density, decomposition temperature, and specific heat. The $x$-axis represents reference values and the $y$-axis represents model predictions. GATE predictions (top) are benchmarked against single-task learning (STL) models (bottom). Pearson correlation coefficients ($r$) and mean absolute errors (MAE) are shown.
    \textbf{c.} Overall comparison of GATE and STL performance across 34 physicochemical properties. Bars indicate Pearson correlation coefficients for validation and OOD sets, highlighting GATE’s stronger generalization across diverse properties.}
    \label{fig:ood_benchmark}
\end{figure}

Most real-world materials discovery problems demand compounds with unusual combinations of properties—often extreme or unconventional values that lie outside existing data distributions. Such out-of-distribution (OOD) regions represent where AI models are most practically tested in real materials development, as they correspond to property domains rarely captured in training data yet critical for innovation.

To assess GATE’s performance in these realistic OOD scenarios, we used the discovery of immersion-coolant candidates as a representative testbed. These candidates lie far beyond the training domain but exemplify the type of chemical space often encountered in emerging materials discovery problems. The goal here is not limited to immersion cooling itself, but to benchmark how a generalizable model like GATE performs when directly applied—without retraining or adaptation—to a truly OOD problem.

Building on the adapted specifications summarized in Table~\ref{tab:ocp} (right), we grouped the screening requirements into five categories:
\begin{itemize}
    \item \textbf{Thermal}: boiling point $>$ 150~\textcelsius{}, melting point $<$ –30~\textcelsius{}, critical temperature $>$ 155~\textcelsius{}, and decomposition temperature $>$ 140~\textcelsius{}
    \item \textbf{Pressure}: vapor pressure $<$ 0.8~kPa
    \item \textbf{Figures of merit}: dynamic viscosity $<$ 0.015~N·s/m\textsuperscript{2} and density $<$ 2,000~kg/m\textsuperscript{3}
    \item \textbf{Electrical}: dielectric constant $\leq$ 6 at 1~kHz
    \item \textbf{Structural}: no S, F, Cl, Br, I, or aromatic rings
\end{itemize}

These groups mirror the adapted criteria used throughout this study and the color scale in Fig.~\ref{fig:ood_benchmark}a, where each data point is colored by the number of criteria groups it satisfies.

We evaluated a set of dielectric liquids that share similar physicochemical characteristics yet differ in how they were selected. The first group comprises the four GATE-discovered candidates identified earlier through the high-throughput screening workflow, together with a commercial immersion fluid, DOWSIL\textsuperscript{TM} ICL-1000. The second group consists of 13 additional dielectric liquids that are non-halogenated, liquid at room temperature, and readily purchasable. These additional liquids satisfy all non-thermal groups but not the full thermal requirements (–30~\textcelsius{}~$\leq~T~\leq$~150~\textcelsius{}), corresponding to the thermal group defined above—encompassing thresholds for boiling point, melting point, critical temperature, and decomposition temperature. Despite these differences in selection, all evaluated liquids are electrically insulating and lie well outside the GATE training domain, representing out-of-distribution (OOD) materials as shown in Fig.~\ref{fig:ood_benchmark}a (stars: GATE candidates, square: commercial fluid, triangles: other dielectric liquids). Detailed information on all evaluated liquids is summarized in Supplementary Table 3.

Among the 34 physicochemical properties supported by GATE, five—boiling point, flash point, density, decomposition temperature, and specific heat—had at least two independent experimental or literature reference values available for comparison. Reference data were collected from vendor technical or safety data sheets (Sigma-Aldrich, Alfa Chemistry, TCI) and literature sources, and supplemented by our own measurements of decomposition temperature (TGA) and specific heat (DSC). These properties therefore represent all cases where quantitative benchmarking was possible under OOD conditions.

Across the five experimentally or literature-accessible properties, GATE maintained strong predictive accuracy under OOD conditions (Fig.~\ref{fig:ood_benchmark}b). Pearson correlation coefficients between predictions and reference values ranged from 0.65 (specific heat) to 0.96 (density), exceeding those of STL models for all benchmarked properties. Notably, GATE preserved meaningful correlations even in the most challenging cases—such as decomposition temperature and specific heat, the latter having only 387 training data points—where STL failed to generalize beyond its training domain (see Fig.~\ref{fig:ood_benchmark}c).

GATE achieved lower MAEs than STL, with relative reductions ranging from –74.4\% (density) to –23.8\% (specific heat). As designed, GATE transfers knowledge across correlated properties through geometric alignment of task-specific manifolds, allowing information learned from one task to reinforce others. This cross-property transfer stabilizes learning in data-scarce and extrapolative regimes, enabling robust prediction even under OOD conditions.

In addition to its strong performance under OOD conditions, GATE also outperformed STL within the training domain. Among the 34 physicochemical properties, GATE achieved higher Pearson correlations than STL for 23 properties (68\%). Notably, 11 properties where STL slightly outperformed GATE, the performance gap was minimal ($\Delta r = 0.009-0.030$), whereas for those where GATE excelled, the gains were substantially larger ($\Delta r = 0.011-0.252$). While STL is optimized for individual properties, GATE jointly balances many correlated targets. Minor trade-offs are expected when optimizing multiple objectives simultaneously, yet this holistic approach yields a more consistent and reliable predictor even within well-sampled regions.

\section*{Discussion}

This study demonstrates how GATE advances generalizable AI for materials discovery by validating its applicability in a real-world challenge. Unlike conventional single-task learning (STL), which treats each property in isolation, GATE aligns property-specific latent spaces through geometric transfer. This enables robust learning even for data-scarce or heterogeneous properties, yielding a more holistic understanding of structure–property relationships. As a result, GATE consistently outperforms STL across benchmarks.

A key concept revealed by this study is the \textit{disjoint-property bias}—the systematic tendency of independently trained single-task models to over-retain false positives when multiple design criteria are applied simultaneously. Such compounds can appear promising \textit{in silico} but fail to meet all requirements experimentally. In contrast, GATE learns the underlying correlations among properties within a shared representation, mitigating this bias and providing greater confidence that screened candidates will satisfy multiple constraints at once.

To assess GATE under more realistic conditions, we benchmarked its predictions against experimental measurements and literature values in out-of-distribution (OOD) regimes, where promising materials often lie beyond the boundaries of existing data. Across the five measurable properties, GATE consistently outperformed STL, reducing MAEs by up to –74.4\%. Notably, for specific heat—the property with the fewest training samples (387)—GATE achieved a 242\% higher Pearson correlation with experimental and literature values compared to STL. These results demonstrate that GATE maintains reliable predictive performance even in data-scarce regions, highlighting its robustness in extrapolative, real-world discovery scenarios. By leveraging cross-property correlations, GATE mitigates the dual challenges of data scarcity and \textit{disjoint-property bias}, providing more trustworthy predictions for practical materials development.

GATE’s generalizability arises from its architecture and training philosophy. The model jointly learns 34 physicochemical properties encompassing thermal, electrical, mechanical, and interfacial domains. These properties were selected through extensive consultation with domain experts from diverse materials development fields—including polymers, electronic materials, batteries, displays, cosmetics, and agricultural chemicals, as well as functional additives—to capture the key physicochemical parameters most critical to real-world materials design. This cross-domain learning enables knowledge transfer among correlated properties, allowing the model to predict a wide range of material behaviors without retraining. Building on this foundation, GATE was practically validated through the discovery of immersion cooling fluids for data centers—a stringent, multi-criteria problem defined by the OCP guidelines. Without any model reconfiguration, GATE screened billions of compounds and identified 92,861 candidates meeting these specifications, several of which exhibited experimentally or literarily verified properties comparable to a commercial coolant. Beyond this proof-of-concept, GATE establishes a unified and ready-to-use foundation for accelerating materials discovery across domains, illustrating that its applicability extends far beyond a single case.

Finally, we highlight broader challenges that must be addressed for AI-driven materials discovery to mature. Inconsistent measurements, impurities, and loss of contextual metadata introduce substantial noise into chemical databases, limiting the achievable accuracy of any model. Tackling these systemic issues will be essential for AI systems like GATE to fully realize their potential in guiding reliable and scalable materials innovation (see Lessons Learned).

\section*{Lessons learned}
While GATE provides broadly consistent and practically useful predictions, some discrepancies remain between internal validation and experimental benchmarking. These differences primarily reflect limitations in data quality rather than the model itself.  

Chemical property datasets—even widely used ones—often contain contaminated or inconsistent entries. Measurements are frequently conducted on impure or mixed samples, leading to inherent variability in reported values. For the same property, different conventions—such as whether to report the initial, onset, or inflection of a measured signal—can yield divergent results that are rarely harmonized. During large-scale aggregation, essential contextual metadata such as pressure, temperature, or inequality markers (e.g., boiling point 61~\textcelsius{} “at 0.03~bar,” flash point “$>$100~\textcelsius{}”) are frequently lost. Such errors propagate through public databases and even into vendor safety data sheets (SDS), amplifying noise across the ecosystem.  

A particularly pervasive example involves thermogravimetric analysis (TGA). Although TGA is designed to measure decomposition temperatures of solid samples, it is often misapplied to liquids, where mass loss can also arise from evaporation or boiling rather than true chemical breakdown. Because the instrument cannot distinguish these processes, the resulting “decomposition temperature” may reflect phase change rather than decomposition. Upon examining widely used public datasets, we found that a substantial fraction of decomposition-temperature entries corresponded to liquid samples, and many of these values were nearly identical to their reported boiling points. This systematic ambiguity indicates that a large portion of publicly available “decomposition temperature” data may in fact represent thermal volatilization rather than genuine decomposition.  

These problems are not unique to GATE but are endemic to chemical informatics datasets. They underscore the need for curated, standardized, and reproducible measurements that preserve full experimental context. Addressing such data contamination at its source is essential to ensure that AI-driven materials discovery rests on robust and trustworthy experimental foundations. Looking forward, we believe that establishing standardized, context-rich measurements will be critical for enabling the next generation of reliable and transferable AI models in materials science.

\section*{Methods}

\subsection*{\textit{Disjoint-property bias}: mathematical formulation \& proof}

When independently trained single-property models are combined for multi-criteria screening, they implicitly assume statistical independence among properties. This independence assumption inflates the estimated probability that a candidate simultaneously satisfies multiple property thresholds. We term this effect the \textit{disjoint-property bias}. Its origin and mitigation are formalized below.

\begin{itemize}

\item \textbf{Setup and notation.} 
Let $(X, Y)$ denote two real-valued properties of a candidate, with thresholds $t_1$ and $t_2$. Define
\[
A^{\uparrow} := \{X > t_1\}, 
\quad 
B^{\uparrow} := \{Y > t_2\}, 
\quad 
B^{\downarrow} := \{Y < t_2\}.
\]
Under the independence assumption, the joint satisfaction probabilities are approximated as
\[
\hat{P}(A^{\uparrow} \cap B^{\uparrow}) = P(A^{\uparrow})\,P(B^{\uparrow}), 
\qquad
\hat{P}(A^{\uparrow} \cap B^{\downarrow}) = P(A^{\uparrow})\,P(B^{\downarrow}).
\]

\item \textbf{Indicator covariance formulation.} 
For any events $U, V$,
\[
P(U \cap V) = P(U)\,P(V) + \mathrm{Cov}(\mathbf{1}_U, \mathbf{1}_V).
\]
Therefore, independence \emph{overestimates} the joint whenever $\mathrm{Cov}(\mathbf{1}_U, \mathbf{1}_V) < 0$. The sign of this covariance depends on (i) whether $\mathbf{1}_U, \mathbf{1}_V$ are increasing or decreasing in $X, Y$, and (ii) the dependence between $X$ and $Y$. We use the notion of quadrant dependence: positively quadrant dependent (PQD) means $P(X>u, Y>v) \ge P(X>u)\,P(Y>v)$, while negatively quadrant dependent (NQD) means $P(X>u, Y>v) \le P(X>u)\,P(Y>v)$.

\item \textbf{Case I — Antagonistic properties (negative dependence, same-direction thresholds).}

\textbf{\textit{Claim:}}
\[
P(A^{\uparrow} \cap B^{\uparrow}) = P(A^{\uparrow})\,P(B^{\uparrow})+ \mathrm{Cov}(\mathbf{1}_{A^{\uparrow}}, \mathbf{1}_{B^{\uparrow}})\le P(A^{\uparrow})\,P(B^{\uparrow}), 
\quad \text{equivalently } 
\mathrm{Cov}(\mathbf{1}_{A^{\uparrow}}, \mathbf{1}_{B^{\uparrow}}) \le 0.
\]

\textbf{\textit{Reasoning:}}
Both $\mathbf{1}_{A^{\uparrow}}$ and $\mathbf{1}_{B^{\uparrow}}$ are increasing functions. For NQD variables, $\mathrm{Cov}(f(X), g(Y)) \le 0$ for any non-decreasing $f,g$ (Lehmann, 1966)\cite{Lehmann1966}. Applying this with $f(X)=\mathbf{1}_{A^{\uparrow}}$ and $g(Y)=\mathbf{1}_{B^{\uparrow}}$ yields
\[
P(X>t_1, Y>t_2) \le P(X>t_1)\,P(Y>t_2),
\]
thus independence overestimates the joint feasibility.

\textbf{\textit{Gaussian example:}}
For a bivariate normal with correlation $\rho \le 0$, $P(X>t_1, Y>t_2)$ increases with $\rho$, hence
\[
P(X>t_1, Y>t_2)\mid_{\rho\le0} \;\le\; P(X>t_1, Y>t_2)\mid_{\rho=0} = P(A^{\uparrow})\,P(B^{\uparrow}).
\]

\item \textbf{Case II — Co-varying properties (positive dependence, opposing thresholds).}

\textbf{\textit{Claim:}}
\[
P(A^{\uparrow} \cap B^{\downarrow}) = P(A^{\uparrow})\,P(B^{\downarrow}) + \mathrm{Cov}(\mathbf{1}_{A^{\uparrow}}, \mathbf{1}_{B^{\downarrow}}) \le P(A^{\uparrow})\,P(B^{\downarrow}), 
\quad \text{equivalently } 
\mathrm{Cov}(\mathbf{1}_{A^{\uparrow}}, \mathbf{1}_{B^{\downarrow}}) \le 0.
\]

\textbf{\textit{Reasoning:}}
Here $\mathbf{1}_{A^{\uparrow}}$ is increasing in $X$, while $\mathbf{1}_{B^{\downarrow}}$ is decreasing in $Y$. For PQD variables, an increasing–decreasing pair yields $\mathrm{Cov}(f(X), g(Y)) \le 0$ because $(X, -Y)$ is NQD and both $f(X)$ and $g(-Y)$ become non-decreasing (Lehmann, 1966)\cite{Lehmann1966}. Hence
\[
P(X>t_1, Y<t_2) \le P(X>t_1)\,P(Y<t_2).
\]

\textbf{\textit{Gaussian example:}}
For a bivariate normal with correlation $\rho \ge 0$, $P(X>t_1, Y<t_2)$ decreases with $\rho$, thus
\[
P(X>t_1, Y<t_2)\mid_{\rho\ge0} \;\le\; P(X>t_1, Y<t_2)\mid_{\rho=0} = P(A^{\uparrow})\,P(B^{\downarrow}).
\]

\item \textbf{Geometric mitigation via GATE.}
GATE alleviates the bias by learning cross-property dependencies in a shared latent space $z = E(x) \in \mathbb{R}^d$. Instead of independent mappings, GATE jointly optimizes
\[
L = \sum_i L_i(f_i(x), y_i) + \lambda \sum_{i<j} D_{\text{align}}(z_i, z_j),
\]
where $D_{\text{align}}$ penalizes geometric misalignment of property manifolds. This 
(1) recovers correlation sign and magnitude between properties;
(2) encodes feasible trade-offs between antagonistic/co-varying properties;
(3) corrects the inflated independence product toward a calibrated joint probability $P_{\text{GATE}}(\text{all})\approx P_{\text{true}}(\text{all})$.

\item \textbf{Consequence.}
In both trade-off archetypes—(i) negative dependence with same-direction thresholds and (ii) positive dependence with opposing thresholds—we have:
\[
P(\text{joint}) \le P(\text{independent product}),
\]
which shows independence-based multi-criteria screening systematically overestimates feasibility and inflates false positives. By geometrically aligning property manifolds in a shared latent space, GATE embeds inter-property correlations and provides calibrated multi-criteria predictions.

\end{itemize}

\subsection*{Virtual library construction}
The virtual screening library was built by combining commercially available compounds with virtually synthesized derivatives (Fig.~\ref{fig:workflow}b). We obtained 42 million compounds from the Mcule database\cite{mcule} (licensed access) and 73,562 compounds from Sigma–Aldrich (via PubChem\cite{pubchem}), resulting in 41 million unique structures after deduplication.

To expand this space, we selected single-step synthetic transformations in consultation with domain experts, focusing on reactions that are both reliable and feasible under mild conditions. Specifically, we applied Williamson ether synthesis (alcohol + alkyl chloride → ether) and esterification (alcohol + carboxylic acid → ester). Prior to virtual synthesis, structure-based screening was applied to remove compounds containing halogens (except chlorine), sulfur, aromatic rings, or unstable substructures (see next section). This process yielded 347,928 suitable reactants, including 44,707 alcohols, 10,719 chlorides, and 46,016 carboxylic acids.

Crossing these sets generated approximately 0.48 billion ethers and 2.06 billion esters, producing a total of 2.54 billion virtual compounds. Together with the 41 million purchasable molecules, this collection formed the input pool for subsequent AI-based screening.

\subsection*{Structure-based screening}
Prior to virtual synthesis, compounds containing structural motifs unsuitable for immersion cooling fluids were removed (Fig.~\ref{fig:workflow}c). Specifically, molecules bearing halogens other than chlorine (F, Br, I) were excluded because of their association with ozone depletion, toxicity, and regulatory restrictions. Compounds containing sulfur were also eliminated due to their corrosivity, unpleasant odor, and tendency to impart yellow–brown coloration. Aromatic rings were filtered out, as they are often linked to odor issues and, in some cases, carcinogenicity. In addition, molecules with strained rings containing four or fewer atoms were discarded owing to their inherent instability.

Chlorine was retained at this stage to allow alkyl chlorides as reactants for the Williamson ether synthesis used in library expansion. However, following virtual synthesis, all chlorine-containing materials were subsequently removed to ensure that final candidates satisfied environmental and safety constraints.

This structure-based screening step ensured that subsequent screening focused on candidates with environmentally benign, chemically stable, and practically acceptable molecular backbones.

\subsection*{AI-based screening strategy}

For GATE predictions, we employed an ensemble of eight models obtained from 8-fold cross-validation, averaging their outputs to improve robustness. While this improves accuracy, it also increases computational demand: on a single NVIDIA A40 GPU, GATE processes 98 molecules per second, which would require approximately 10 months to evaluate all 2.58 billion candidates.

To address this limitation, we developed a surrogate model that leverages the manifold representations learned by GATE (Supplementary Fig.~3). Rather than relying solely on structural descriptors, the surrogate concatenates the precomputed manifold vectors of two reactants (stored as a lookup table) and predicts the properties of their product. This design enables the surrogate to approximate GATE outputs with minimal accuracy loss, owing to its use of GATE’s chemically informed manifold.

The surrogate was trained on 0.5\% of all possible product combinations (90/10 train/validation split), achieving throughput of 3,000 molecules per second on the same A40 GPU. Importantly, Pearson correlations with GATE predictions exceeded 0.9 across most properties (Supplementary Table~1), demonstrating that the surrogate preserved the predictive patterns of GATE with only minimal loss in accuracy. Because surrogate predictions are inherently less accurate than GATE, we applied them with relaxed thresholds (Supplementary Table~2), ensuring that viable candidates were retained for final evaluation.

This two-stage strategy—surrogate pre-screening followed by GATE-based refinement—effectively reduced the computational cost. Direct GATE evaluation of all $n \times m$ product combinations would have been prohibitive, but by computing GATE manifolds for only $n+m$ reactants and then using the surrogate for combinatorial expansion, the effective burden was reduced from $O(nm)$ heavy GATE evaluations to $O(n+m)$ GATE runs plus lightweight surrogate inference. This made large-scale screening practical while maintaining predictive reliability.

\subsection*{Synthesis of candidate 4}
2-Butyl-1-octanol (16 mmol), 2-methyl-2-pentenoic acid (16 mmol) and 1-methyl-3-(4-sulfobutyl)imidazolium hydrogen sulfate (0.8 mmol) were dissolved in tetrahydrofuran (25 mL) and refluxed for 24 h. After the completion of reaction, mixture dissolved in chloroform was washed with distilled water, and then organic solvent layer was extracted. Candidate 4 was obtained from removing the remaining chloroform using rotary evaporator (yield 87\%).

To verify that the desired compound had been successfully synthesized, structural confirmation was performed using \textsuperscript{1}H NMR spectroscopy. The full spectral data and corresponding spectra are provided in Supplementary Fig.~4.

\subsection*{Thermal analysis}
TGA was carried out on a PerkinElmer TGA4000 under nitrogen, heating samples from 40 to 700 \textcelsius{} at a rate of 10 \textcelsius{}/min. Specific heat capacity was measured using a DSC (Discovery DSC25) following the ASTM E-1269 protocol, with sapphire as the reference material. Samples were analyzed in modulated mode ($\pm 1$ \textcelsius{} amplitude, 120 s period) from $-$40 to 80 \textcelsius{} at a heating rate of 2 \textcelsius{}/min under nitrogen atmosphere.

\subsection*{Architecture of GATE}
As shown in Fig.~\ref{fig:gate}a, the GATE model consists of five components: an embedding network, encoder, transfer network, inverse transfer network, and task-specific prediction head. Molecular structures are represented as graphs and encoded using a Directed Message Passing Neural Network (DMPNN, depth = 2) followed by a bottleneck autoencoder. Task-specific transfer and inverse transfer networks (multi-layer perceptrons) align the latent representations of different physicochemical properties within a shared Riemannian manifold. Perturbation-based distance regularization is applied to preserve the local geometry of the manifold across tasks.

The overall training objective combines five loss terms:
\[
L_{\mathrm{tot}} = L_{\mathrm{reg}} + \alpha L_{\mathrm{auto}} + \beta L_{\mathrm{cons}} + \gamma L_{\mathrm{map}} + \delta L_{\mathrm{dis}},
\]
with all weights set to 1. Their definitions are as follows:

\begin{itemize}
    \item \textbf{Regression loss} ($L_{\mathrm{reg}}$): mean squared error between the predicted property $\hat{y}_i$ and its reference value $y_i$,
    \[
    L_{\mathrm{reg}} = \frac{1}{N} \sum_{i=1}^N (y_i - \hat{y}_i)^2.
    \]

    \item \textbf{Autoencoder loss} ($L_{\mathrm{auto}}$): reconstruction error between the original latent vector $z_i$ and the one reconstructed by the inverse transfer network $\hat{z}_i$,
    \[
    L_{\mathrm{auto}} = \frac{1}{N} \sum_{i=1}^N \| z_i - \hat{z}_i \|^2.
    \]

    \item \textbf{Consistency loss} ($L_{\mathrm{cons}}$): enforces alignment between latent embeddings of the same molecule when mapped through different task-specific transfer networks,  
    \[
    L_{\mathrm{cons}} = \frac{1}{N} \sum_{i=1}^N \| m^{(s)}_i - m^{(t)}_i \|^2,
    \]
    where $m^{(s)}_i$ and $m^{(t)}_i$ denote the transferred latent vectors from source and target tasks.

    \item \textbf{Mapping loss} ($L_{\mathrm{map}}$): ensures that a source latent vector transformed into the target coordinate system remains predictive of the target property $y_i$,  
    \[
    L_{\mathrm{map}} = \frac{1}{N} \sum_{i=1}^N (y_i - \hat{y}'_i)^2,
    \]
    where $\hat{y}'_i$ is the property value predicted from the mapped latent vector.

    \item \textbf{Distance loss} ($L_{\mathrm{dis}}$): regularizes manifold geometry by matching pairwise distances between pivot points $x_i$ and their perturbed versions $x_{ij}^{\mathrm{pert}}$ across tasks,  
    \[
    L_{\mathrm{dis}} = \frac{1}{NM} \sum_{i=1}^N \sum_{j=1}^M \| (m_i - m_{ij}^{\mathrm{pert}})^{(s)} - (m_i - m_{ij}^{\mathrm{pert}})^{(t)} \|^2.
    \]
\end{itemize}

Models were trained for 600 epochs with a batch size of 512 using AdamW (learning rate = $5\times10^{-5}$) and 10 perturbations per sample.

\subsection*{Architecture of STL}
For comparison, we implemented a conventional single-task learning (STL) model. STL uses the same backbone encoder (DMPNN) and MLP regression head as GATE but trains each physicochemical property independently. Unlike GATE, no transfer or alignment modules are included, and the model minimizes only the regression loss $L_{\mathrm{reg}}$.

\subsection*{Architecture of surrogate model}
The surrogate model was designed as a lightweight approximation of GATE to rapidly predict properties of products generated from combinatorial reactant pairs, for which direct GATE inference would be computationally prohibitive. Unlike GATE, which encodes each candidate molecule individually, the surrogate operates on \emph{precomputed} reactant embeddings derived from the GATE manifold. Specifically, manifold vectors of all reactants are first computed using GATE and stored in a lookup table. For a product candidate formed by reactants $r_1$ and $r_2$, the input is constructed by concatenating their GATE manifold embeddings,
\[
\mathbf{x} = \mathrm{concat}\!\big(\mathbf{e}_{r_1},\, \mathbf{e}_{r_2}\big) \in \mathbb{R}^{2d},
\]
thereby eliminating the need for per-product GATE inference.

The surrogate model adopts a shared-bottom, multi-head MLP architecture. A common trunk processes the input through two hidden layers (64 and 50 units with LeakyReLU activations), followed by task-specific heads. Each head consists of a three-layer tower (50–32–16 units with LeakyReLU activations) terminating in a scalar output. Ten immersion-relevant physicochemical properties were jointly predicted: boiling point, melting point, flash point, critical temperature, decomposition temperature, specific heat capacity, vapor pressure, dynamic viscosity, density, and dielectric constant.

The training objective was the unweighted sum of task-wise mean squared errors:
\[
L_{\mathrm{surr}} = \sum_{t \in \mathcal{T}} \frac{1}{N}\sum_{i=1}^{N} \left(y^{(t)}_i - \hat{y}^{(t)}_i\right)^2,
\]
where $y^{(t)}_i$ is the GATE-predicted value for property $t$ of product $i$, and $\hat{y}^{(t)}_i$ is the surrogate prediction. Here, $\mathcal{T}$ denotes the set of ten physicochemical properties predicted by the surrogate (boiling point, melting point, flash point, critical temperature, decomposition temperature, specific heat capacity, vapor pressure, dynamic viscosity, density, and dielectric constant).

Models were trained with Adam optimization and a batch size of 256. All hidden layers used LeakyReLU activation with negative slope $0.01$. This precompute-and-combine strategy reduces the number of expensive GATE evaluations from $O(nm)$ for $n\times m$ product pairs to $O(n{+}m)$ for individual reactants, while retaining high correlation with GATE outputs across properties. The predictive performance of the surrogate relative to GATE is summarized in Supplementary Table~1.

\section*{Data availability}
The data that support the findings of this study are available from the corresponding author upon reasonable request.

\section*{Code availability}
The GATE model is deployed as part of the \textit{EXAONE Chemical Agent} platform developed by LG AI Research. 
A hosted version of the model is accessible through the \textit{EXAONE Showroom} (\href{https://showroom.exaone.ai}{https://showroom.exaone.ai}) upon reasonable request for a trial account. 
The publicly available service may be periodically updated in its architecture and parameters, 
which could result in minor differences from the predictions reported in this paper.

\bibliography{main}

\section*{Author contributions statement}
H.K., D.W.J., C.P., and S.H. conceived the study and contributed to the overall experimental design and data analysis. H.K. carried out the AI-based materials screening. W.J.L., H.E.L. and J.H.L. conducted the materials synthesis and property characterization. H.K., D.W.J., S.M.K., S.L., S.Y. and C.L. contributed to the development of the GATE model. H.K. wrote the manuscript with input from all authors. S.H.C. supervised the material synthesis and experimental measurements, and S.H. and W.L. oversaw the overall project. All authors discussed the results and approved the final manuscript.

\section*{Competing interests}
H.K., D.W.J., C.P., R.H., S.M.K., S.L., S.Y., C.L., S.H., and W.L. are employees of LG AI Research, which funded this study.
W.J.L., H.E.L., J.H.L., and S.H.C. are affiliated with Kyonggi University and collaborated through an industry–academia research program supported by LG AI Research.
The authors declare no other competing interests.

\end{document}